\documentclass[11pt,a4paper]{article}
\usepackage{times,latexsym}
\usepackage{url}
\usepackage[T1]{fontenc}

\usepackage[acceptedWithA]{tacl2021v1}

\usepackage{xspace,mfirstuc,tabulary}

\newif\iftaclinstructions
\taclinstructionsfalse 
\iftaclinstructions
\renewcommand{\confidential}{}
\renewcommand{\anonsubtext}{(No author info supplied here, for consistency with
TACL-submission anonymization requirements)}
\newcommand{\instr}
\fi

\iftaclpubformat 

\else

\fi

\usepackage{graphicx}
\usepackage{soul} 

\usepackage[strict]{changepage} 
\usepackage{framed} 
\definecolor{formalshade}{rgb}{0.95,0.95,1}

\usepackage{microtype}
\usepackage{amssymb}
\usepackage{multirow}
\usepackage{hyperref}
\usepackage{booktabs}
\usepackage{colortbl}
\usepackage{algorithm}
\usepackage{bm}
\usepackage[noend]{algpseudocode}

\usepackage[hang,flushmargin]{footmisc}  
\newcommand{\LN}{\linebreak\noindent}    
\newcommand{\ComFact}{$\mathcal{C}$omFact}
\newcommand{\Cicero}{$\mathcal{C}$icero}
\newcommand{\Reflect}{$\mathcal{R}$eflect}
\newcommand{\ConvoSense}{$\mathbb{C}$onvo$\mathbb{S}$ense}
\newcommand{\HumanGen}{$\mathbb{H}$uman$\mathbb{G}$en}

\title{ConvoSense: Overcoming Monotonous \\ Commonsense Inferences for Conversational AI}

\author{
  Sarah E. Finch \\
  Department of Computer Science \\
  Emory University \\
  Atlanta, GA, USA \\
  \texttt{sfillwo@emory.edu} \\
  \And
  Jinho D. Choi \\
  Department of Computer Science \\
  Emory University \\
  Atlanta, GA, USA \\
  \texttt{jinho.choi@emory.edu}
}

\date{}

\begin{document}
\maketitle

\begin{abstract}

Mastering commonsense understanding and reasoning is a pivotal skill essential for conducting engaging conversations. 
While there have been several attempts to create datasets that facilitate commonsense inferences in dialogue contexts, existing datasets tend to lack in-depth details, restate information already present in the conversation, and often fail to capture the multifaceted nature of commonsense reasoning.
In response to these limitations, we compile a new synthetic dataset for commonsense reasoning in dialogue contexts using GPT, \ConvoSense, that boasts greater contextual novelty, offers a higher volume of inferences per example, and substantially enriches the detail conveyed by the inferences.
Our dataset contains over 500,000 inferences across 12,000 dialogues with 10 popular inference types, which empowers the training of generative commonsense models for dialogue that are superior in producing plausible inferences with high novelty when compared to models trained on the previous datasets. 
To the best of our knowledge, \ConvoSense\ is the first of its kind to provide such a multitude of novel inferences at such a large scale.

\end{abstract}
\section{Introduction}

Effective dialogue is accomplished by a profound grasp of language and a thorough comprehension of the world.
Such comprehension is crucial to the construction of responses that are pertinent, coherent, and captivating within an ongoing dialogue.
A pivotal element of this worldview is commonsense: self-evident information that is universally acknowledged among humans \cite{clark:91}.


Over time, there has been a concerted endeavor to create datasets that facilitate commonsense reasoning. 
Early work, such as the widely recognized ConceptNet \cite{speer:17}, focused predominantly on physical commonsense related to entities.\LN
Lately, efforts have shifted toward building  datasets encompassing social- and event-based commonsense, such as ATOMIC \cite{hwang:21}. 
This new wave of datasets targets complex human concepts, including emotions, desires, and motivations.


As human conversations largely revolve around sharing personal experiences and life events \cite{fillwock:18, mitsuda:19}, it is critical for virtual agents to possess a robust understanding of human experiences to conduct effective dialogue. 
Datasets such as ATOMIC hold promise as they provide insights directly relevant to human experience; however, a drawback lies in their lack of contextual awareness as they hinge on isolated, concise phrases for commonsense inferences.
This limitation poses challenges for dialogue-oriented tasks because utterances should not be viewed in isolation but must be interpreted within their context \cite{pan:19, jin:22}.


Several initiatives have recently aimed to curate commonsense inferences tailored for dialogue contexts \cite{gao:22, ghosal:22, zhou:22a}.
However, a trade-off currently exists between the breadth of inference types covered and the scope of dialogue contexts encompassed within these existing datasets. 
While some datasets cover a wide range of relations, they are limited to a small number of dialogues \cite{gao:22}, whereas others capture a large number of dialogues but on a limited set of relations \cite{ghosal:22}.

\begin{table*}[htb!]
\centering
\small
\resizebox{\textwidth}{!}{
\begin{tabular}{c|c|p{35em}|c|c|c}
\toprule
\bf Type & \bf Label(s) & \multicolumn{1}{c|}{\bf Definition(s)} & \bf COM & \bf CIC & \bf REF \\
\midrule
\multirow{3}{*}{Subsequent} & isBefore  & What could happen after this? [2]  & \multirow{3}{*}{*} & \multirow{3}{*}{22K} & \multirow{3}{*}{600} \\
 & Subsequent- &  What subsequent event happens or could happen following the Target? [3]  &  &  &  \\
 & Events &  What might happen after? [4] &  &  &  \\
\midrule
\multirow{2}{*}{Antecedent} & \multirow{2}{*}{isAfter} & What could have happened before this? [2]  & \multirow{2}{*}{*} &  & \multirow{2}{*}{600} \\
 &  &  What might have happened before? [4] &  &  &  \\
\midrule
\multirow{2}{*}{Cause} &  xReason  & What could be the cause of this event? [2]  & \multirow{2}{*}{80} & \multirow{2}{*}{21K} &  \\
 &  Cause &  What is the event that directly causes or could cause Target? [3] &  &  &  \\
\midrule
\multirow{2}{*}{Prerequisite} & xNeed  & What does X need to do before the event can happen? [1]  & \multirow{2}{*}{1K} & \multirow{2}{*}{10K} &  \\
 &  Prerequisites &  What is or could be the prerequisite of Target? [3]  &  &  &  \\
\midrule
\multirow{2}{*}{Motivation} & xIntent  & Why does X cause the event? [1]  & \multirow{2}{*}{800} & \multirow{2}{*}{12K} &  \\
 &  Motivation &  What is an emotion or basic human drive that motivates or could motivate Target? [3] &  &  &  \\
\midrule
\multirow{2}{*}{Attribute} & \multirow{2}{*}{xAttr} & How would X be described? [1]  & \multirow{2}{*}{400} &  & \multirow{2}{*}{600} \\
 &  &  How would you describe Speaker? [4]  &  &  &  \\
\midrule
\multirow{2}{*}{Reaction} & \multirow{2}{*}{xReact} & How does X feel after the event? [1]  & \multirow{2}{*}{300} &  & \multirow{2}{*}{600} \\
 &  &  What is Speaker feeling now? [4] &  &  &  \\
\midrule
\multirow{3}{*}{Reaction$_{o}$} & \multirow{3}{*}{oReact} & How do others feel after the event? [1]  & \multirow{3}{*}{70} & \multirow{3}{*}{6K} & \multirow{3}{*}{600} \\
 &  &  What is the possible emotional reaction of the listener in response to target? [3]  &  &  &  \\
 &  &  What is Responder feeling now? [4] &  &  &  \\
\midrule
Desire & xWant & What would X likely want to do after the event? [1] & 1K &  &  \\
\midrule
Desire$_{o}$ & oWant & What would others likely want to do after the event? [1] & 100 &  &  \\
\midrule
Constituents & HasSubEvent & What is a substep that happens within this event? [2] & 800 &  &  \\
\midrule
Obstacle & HinderedBy & What could obstruct the occurrence of this event? [2] & 200 &  &  \\
\midrule
Effect & Causes & What does this event cause to happen? [2] & 30 &  &  \\
\midrule
Effect$_s$ & xEffect & What effect does the event have on X? [1] & 400 &  &  \\
\midrule
Effect$_o$ & oEffect & What effects does the event have on others? [1] & 90 &  &  \\
\bottomrule
\end{tabular}
}
\caption{The inference types covered in existing commonsense datasets (\textbf{COM}/\textbf{CIC}/\textbf{REF}: the numbers of examples in the \ComFact\ / combined \Cicero\ v1 
\& v2 / \Reflect\ datasets, respectively). Each row denotes a unique type from the existing datasets using definitions from [1] \citet{sap:19}, [2] \citet{hwang:21}, [3] \citet{ghosal:22}, [4] \citet{zhou:22a}. Counts are truncated to the nearest order of magnitude. * indicates the type was included but no human-verified instances of it are present.}
\vspace{-1.5em}
\label{tab:human-gen-datasets-types}
\end{table*}

%
%

In addition, a few challenges can be encountered in these datasets.
For example, the inferences in these datasets are often too succinct and derive only straightforward conclusions with minimal elaboration \cite{gao:22}, which do not convey implicit commonsense.
Some studies instruct annotators to recycle information from the ongoing conversation, undermining the speculative nature of inferences and detracting from the potential of offering fresh insights to enhance dialogue understanding \cite{ghosal:22}.
Moreover, although multiple plausible inferences can be drawn from a single dialogue context, only a few datasets support this multifaceted nature \cite{shen:22}, impeding the development of models capable of generating diverse inferences, and thus, limiting their utility in real applications.


We present \ConvoSense, a commonsense dataset generated by GPT encompassing 10 popular inference types with over 500,000 inferences across 12,000 dialogues (\textsection\ref{sec:convosense}).
Our dataset shows greater contextual novelty and enhanced inference diversity and detail while maintaining exceptional reasonability compared to existing datasets (\textsection\ref{sec:gpt_generation}).
We also explore several strategies to build generative models producing inferences for dialogue contexts (\textsection\ref{sec:model_strategies}).
Our experiments show that models trained on \ConvoSense\ excel in generating plausible inferences with greater detail and novelty, compared to ones trained on existing datasets (\textsection\ref{sec:model-evaluation}).
To the best of our knowledge, this is the first dialogue-based commonsense dataset that not only covers an extensive array of inference types at large-scale but also provides a plethora of diverse, novel inferences tailored to each dialogue context.
Our \ConvoSense\ dataset and inference models can be accessed through our open-source project: {\small\url{https://github.com/emorynlp/ConvoSense}}.

\section{Related Work}
\label{sec:related_work}


Recent works have focused on integrating commonsense into various tasks, including story generation and explanation \cite{guan:20,gabriel:21}, dialogue summarization and explanation \cite{ghosal:21, zhou:21b, kim:22a}, and response generation \cite{li:22, sabour:22, zhou:22b}.
Many of\LN these works rely on existing datasets, such as \LN ConceptNet \cite{li:22, zhou:22b} and ATOMIC \cite{sabour:22}, which only contain single-word or short-phrase premises and conclusions.
Although there are commonsense datasets curated for long dialogue contexts, they tend to be of small size \cite{zhou:22a}, express simple inferences \cite{gao:22}, or copy context from the provided utterances \cite{ghosal:22}.


On the other hand, GPT has recently been used to create a variety of datasets. \citet{kim:22b} and \citet{zhan:23} constructed dyadic dialogue datasets at large-scale, while \citet{west:22} generated commonsense triples in the ATOMIC style \cite{hwang:21}.
However, the ATOMIC-style inferences are not necessarily suitable for dialogue, as they struggle to handle long contexts and often lack depth.
Table~\ref{tab:human-gen-datasets-types} summarizes the inference types in existing dialogue-focused commonsense datasets and mappings of synonymous types among them.
In particular, the following 3 datasets\LN are used for comparisons with our work:





\paragraph{ComFact}

\citet{gao:22} mapped dialogue utterances to reasonable inferences from the existing ATOMIC2020 dataset \cite{hwang:21} by using exact string matching and embedding similarity.
Subsequently, human annotators verified the relevance of the retrieved inferences.



\paragraph{Cicero} 

Human participants were tasked with composing responses to five commonsense questions (e.g., \textit{What is the event that directly causes or could cause Target?}) based on dialogue contexts and explicitly instructed to incorporate information from the preceding or forthcoming utterances.
The first version produced a single inference for each example \cite{ghosal:22}, whereas the second version produced multiple examples of both good and bad inferences \cite{shen:22}.


\paragraph{Reflect}

\citet{zhou:22a} supplied both human-generated commonsense inferences and following utterance responses that could be derived from a specified commonsense inference.
The inferences were collected by instructing human participants to answer a commonsense question, while the next-utterance responses were composed by new human participants who were provided with the dialogue context and one of the human-generated inferences.

\section{Evaluating GPT-generated Inferences}
\label{sec:gpt_generation}

In order to support the development of a large-scale and high coverage commonsense dataset for dialogue that improves upon existing works, we hypothesize that we can leverage large language models (LLMs) to accomplish this task in an efficient and low-cost manner. From initial pilot tests of both closed-source (GPT) and open-sourced LLMs (Vicuna and Llama), we find that GPT provides greater reliability in following specific instructions and produces commonsense inferences of overall better quality than the open-sourced LLMs. Consequently, we choose to rely on GPT in this work.

\subsection{Prompt Engineering}
\label{ssec:gpt_prompt_engineering}

Prior to crafting the full \ConvoSense\ dataset, we empirically assess GPT's efficacy in generating reasonable and novel commonsense inferences for dialogue.
To mitigate any unintended bias from in-context examples in the GPT prompt, we adopt a zero-shot generation framework.\footnote{\texttt{gpt-turbo-3.5-301} with a temperature setting of $1.0$.}
GPT prompts are refined iteratively to achieve the optimal outcomes.
An example of the final prompt design, specifically tailored for the \texttt{Desire} inference type, is illustrated in Table~\ref{tab:final_prompt}.

During our development process, we observe that the inferences generated from GPT frequently contain detailed and rich information, thus addressing one of the major limitations of existing works. In addition, to encourage novel inferences from GPT, we include the instruction ``Your answers should provide novel information that is not explicitly shared in the conversation." as seen in Table 2. We observe that this instruction helps in reducing the redundancy of the generated inferences to the information already explicitly shared in the dialogue context, thus addressing a second major limitation of existing works.


\begin{table}[htbp!]
\centering\resizebox{\columnwidth}{!}{
\begin{tabular}{>{\columncolor[gray]{0.85}}cp{1.1cm}l} 
\toprule
 & Speaker: & I just finished cleaning up my kitchen and \\
 & & getting the trash out. \\
 & Listener: & I don't envy you. I hate cleaning. \\
 & Speaker: & I'm the other way. I love cleaning, and then \\
\multirow{-5}{*}{\bf C}
 & & seeing my nice clean kitchen afterwards. \\
\midrule
 & Target: & I'm the other way. I love cleaning,  and then \\
 \multirow{-2}{*}{\bf T}
 & & seeing my nice clean kitchen afterwards. \\
\midrule
\bf Q & Question: & What does Speaker want to do next? \\
\midrule
\bf A & Answer: & As a result, Speaker wants ... \\

\midrule
\multicolumn{3}{l}{In a list titled "Answers", generate several likely answers to}\\
\multicolumn{3}{l}{this question for the target expression, keeping the rest of}\\
\multicolumn{3}{l}{the conversation in mind.}\\
\multicolumn{3}{l}{Your answers should provide novel information that is not }\\
\multicolumn{3}{l}{explicitly shared in the conversation.}\\
\bottomrule
\end{tabular}}
\caption{A GPT prompt example for the \texttt{Desire} inference type. Segments are dynamically modified based on the example and inference type, as highlighted in the gray containers (\textbf{C}: dialogue context, \textbf{T}: target utterance, \textbf{Q}: inference question, \textbf{A}: inference answer template).}
\label{tab:final_prompt}
\vspace{-0.5em}
\end{table}

\begin{table*}[t]
    \centering
    \small 
    \resizebox{\textwidth}{!}{
    \begin{tabular}{c|l|l}   
\toprule
\bf Type & \multicolumn{1}{c}{\bf Question} & \multicolumn{1}{c}{\bf Answer Template} \\
\midrule
\cellcolor{gray!20} \bf Subsequent & What might happen after what Speaker just said? & After this, ... \\
Antecedent & What events happened before the situation that Speaker just shared? & Before this, ... \\
\cellcolor{gray!20} \bf Cause & What could have caused the last thing said to happen? & This was caused by... \\
\cellcolor{gray!20} \bf Prerequisite &  What prerequisites are required for the last thing said to occur? & For this to happen, it must be true that... \\
\cellcolor{gray!20} \bf Motivation & What is an emotion or human drive that motivates Speaker based on what they just said? & Speaker is motivated... \\
\cellcolor{gray!20} \bf Attribute & What is a likely characteristic of Speaker based on what they just said? & Speaker is... \\
\cellcolor{gray!20} \bf Reaction & How is Speaker feeling after what they just said? & Speaker feels... \\
\cellcolor{gray!20} \bf Reaction$_o$ &  How does Listener feel because of what Speaker just said? & Listener feels... \\
\cellcolor{gray!20} \bf Desire & What does Speaker want to do next? & As a result, Speaker wants... \\
\cellcolor{gray!20} \bf Desire$_o$ & What will Listener want to do next based on what Speaker just said? & As a result, Listener wants... \\
\cellcolor{gray!20} \bf Constituents & What is a breakdown of the last thing said into a series of required subevents? & This involves... \\
Obstacle & What would cause the last thing said to be untrue or unsuccessful? & This is untrue or unsuccessful if... \\
Effect & What does the last thing said cause to happen? & This causes... \\
Effect$_s$ & How does the last thing said affect Speaker? & This causes Speaker to... \\
Effect$_o$ & How does the last thing said affect Listener? & This causes Listener to... \\
\bottomrule
    \end{tabular}
    }
    \caption{Question and answer prefixes used for generating each inference type from GPT for dialogue contexts. The ten inference types used in our work are represented in gray shading.} 
    \label{tab:gpt-prompts}
\vspace{-1.5em}
\end{table*}

\noindent For the prompt, each inference type is paired with a guiding question and an answer prefix, ensuring uniformity in the generated content for the specific type, which respectively fill the \textit{Inference Question} (\textbf{Q}) and \textit{Inference Answer Template} (\textbf{A}) slots in the prompt.
For every dialogue context, a sequence of utterances in the context is placed in the \textit{Dialogue Context} (\textbf{C}) slot, and its final turn gets duplicated in the \textit{Target Utterance} (\textbf{T}) slot.
Finally, the GPT output, commencing with the header \textit{Answers} and adopting a list-like format with newline separation, is parsed to extract the generated inferences.
Table \ref{tab:gpt-prompts} details the questions and answer prefixes employed for the fifteen identified inference types derived from the previous studies in Table \ref{tab:human-gen-datasets-types}.


\subsection{Evaluation}
\label{sec:gpt_commonsense_metrics}

To evaluate the quality of GPT-generated commonsense inferences for dialogues, we compare their \textbf{reasonability} and \textbf{novelty} against inferences from human datasets.
First, we sample a uniform distribution over inference types for each existing dataset.
For every sample, we then prompt GPT to produce relevant inferences and randomly select one from the generated list.
Finally, two human annotators are presented with the dialogue context, inference question, and both the GPT- and human-generated inferences and asked to categorize them for reasonability and novelty.
For this evaluation, we enlist native English speakers via the Surge AI crowdsourcing platform ({\small\url{www.surgehq.ai}}) by paying them at a rate of \$0.15 per sample with an estimated completion time of 45 seconds.


\paragraph{Reasonability} 

Most prior commonsense datasets assess their inferences based on human-judged reasonability \cite{hwang:21, ghosal:22, shen:22, zhou:22a}. 
An inference is deemed reasonable if it makes sense in, is relevant to, and is consistent with the provided dialogue context. 
We follow \citet{hwang:21}, in which annotators categorize inferences into levels of the truth likelihood: \textit{always/likely}, \textit{sometimes/possible}, \textit{never/farfetched}, or \textit{invalid/nonsense}.


\paragraph{Novelty} 

A key trait of commonsense for dialogue is its role in enhancing dialogue comprehension by providing relevant contextual information. 
While \citet{ghosal:22} gauge creativity in human responses, creativity is not strictly focused on inference novelty. 
In our study, annotators evaluate the extent to which an inference contributes fresh information to the conversation, categorized as: \textit{new \& detailed}, \textit{new \& simple}, and \textit{purely repetitive}.

\vspace{0.5em}
\noindent Since we aim to elicit the natural commonsense understanding learned by each annotator through their life experience in our annotation tasks, we do not provide any training or explicit examples towards what constitutes a “reasonable” or “novel” commonsense inference to avoid artificially polluting their commonsense understanding of the world. Instead, we provide a description of the task with definitions of the different categories. Our instructions are intended to mitigate bias towards trivial inference properties by providing clear definitions of the characteristics under study and emphasizing important aspects to keep in mind, such as ignoring grammar errors unless it made an inference nonsensical. Furthermore, decomposing inference quality into two characteristics allows for their independent evaluation.  We verified through pilots that this approach resulted in reliable and reasonable annotations from our annotators for both tasks.

\begin{figure*}[htb!]
    \centering
    \includegraphics[width=\textwidth]{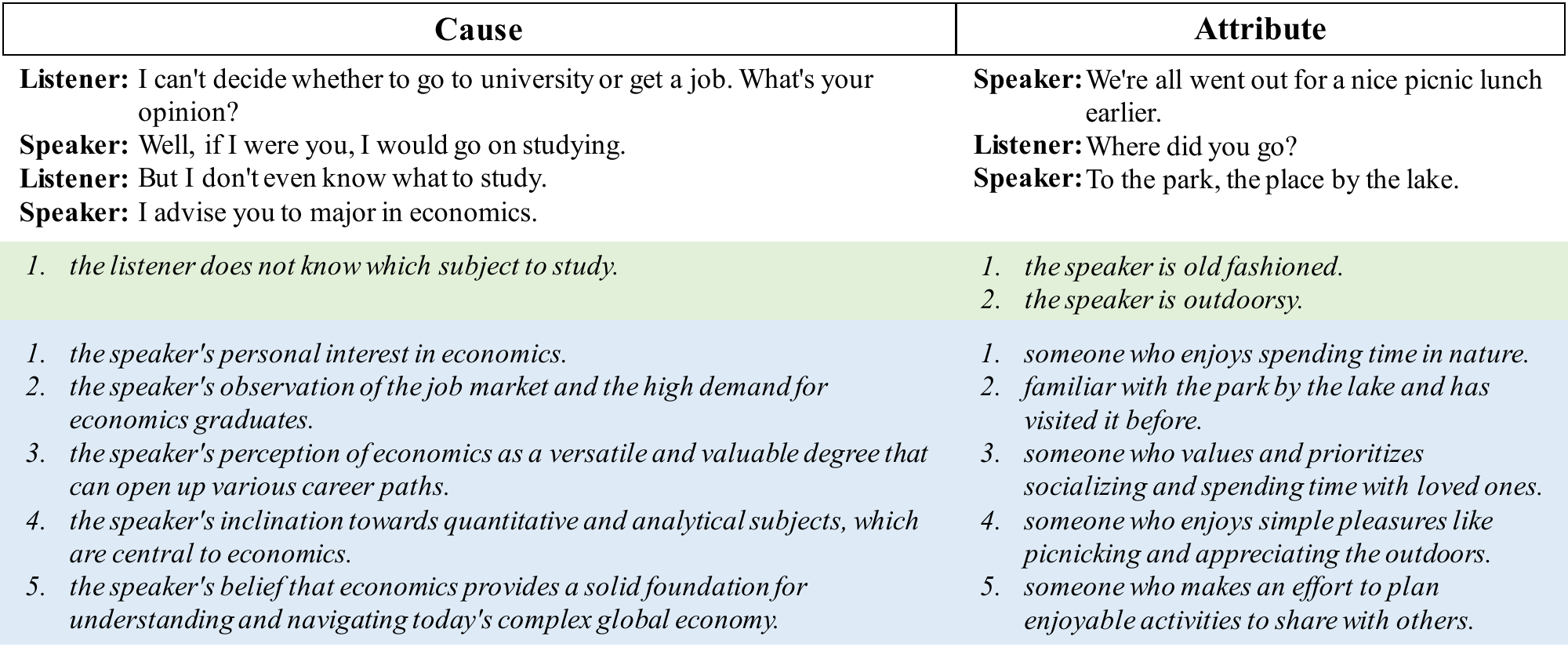}
    \vspace{-1.5em}
    \caption{Cause and Attribute inferences written by humans (top, green) and generated by GPT (bottom, blue).}
    \label{fig:gpt_vs_human_examples}
\vspace{-1em}
\end{figure*}

\subsection{Results}
\label{sec:gpt_commonsense_results}

Following \citet{hwang:21}, the two metrics in Sec.~\ref{sec:gpt_commonsense_metrics} are converted into binary representations.
Thus, labels [\textit{always/likely}, \textit{sometimes/possible}] are categorized as \textit{positive} and [\textit{never/farfetched}, \textit{invalid/nonsense}] are considered \textit{negative} reasonability.
Similarly, [\textit{new \& detailed}, \textit{new \& simple}]\LN are designated as \textit{positive}, and [\textit{purely repetitive}] is classified as \textit{negative} novelty.
This setup, with 300+ annotated samples per dataset, allows us to detect differences of at least 10\% between GPT- and human-generated datasets using McNemar's binary matched-pairs test at 80\% power and a significance level of 0.05, assuming discordance probabilities of 0.24 or lower (compatible with pilots).\footnote{\footnotesize\url{https://homepage.univie.ac.at/robin.ristl/samplesize.php}}
In cases\LN of annotator disagreement, one of the annotators' decisions is randomly selected.
To mitigate the potential noise introduced by this random selection, we repeat the process 100 times and report the average result, only confirming statistical significance when every selection yields a significant result.

Considering the reported quality of the existing\LN datasets and our preliminary assessments of GPT-generated inferences, we expect much higher rates of positive classes than negative ones, resulting in a class imbalance.
To overcome the vulnerability to prevalence skew exhibited by other agreement metrics like Cohen's kappa \cite{jeni:13, wongpakaran:13, quarfoot:16}, Gwet's AC1 inter-annotator agreement metric is chosen \cite{gwet:02}.\footnote{We observe Cohen's kappa of 0.19 and 0.15 for reasonability and novelty, respectively.} Our annotators obtain AC1 values of $0.8$ and $0.6$ for reasonability and novelty, respectively, implying substantial agreement.



\begin{figure}
    \centering
    \includegraphics[width=\columnwidth]{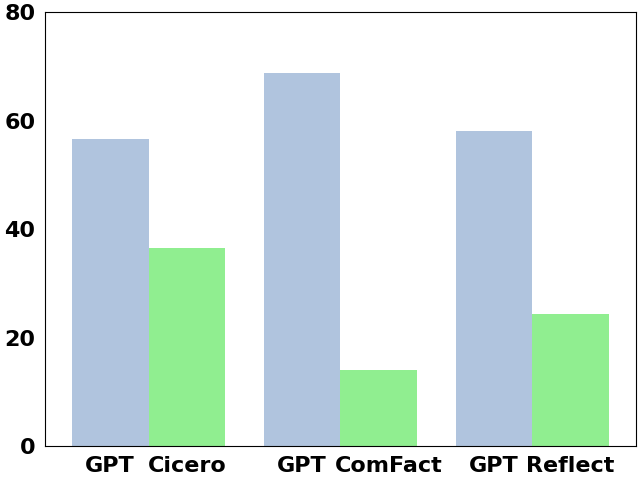}
    \vspace{-1.5em}
    \caption{Average \% of \textit{new \& detailed} inferences out of all positive novelty inferences for each data source.}
    \label{fig:gpt_vs_human_detailedness}
    \vspace{-1em}
\end{figure}

Table~\ref{tab:quality_results} demonstrates that GPT can attain comparable reasonability in its generated inferences as those derived from humans, even exceeding the reasonability of the inferences in \ComFact\ with statistical significance. 
Notably, the results also indicate that GPT \textit{surpasses} the novelty of the human-generated inferences for the majority of the existing datasets. Furthermore, GPT outputs achieve higher detail than that observed from human-generated inferences. Figure \ref{fig:gpt_vs_human_detailedness} shows the percentage of \textit{new \& detailed} inferences out of all positive novelty inferences for each data source, clearly demonstrating the superiority of GPT inferences in terms of their expressed detail. Example inferences from GPT and humans are shown in Figure \ref{fig:gpt_vs_human_examples}.

\begin{table}
\centering
\small{ 
\begin{tabular}{c|l|l|l}
    \toprule
    \bf Dataset & \multicolumn{1}{c|}{\bf R}      & \multicolumn{1}{c|}{\bf N}      & \multicolumn{1}{c}{\bf \#}               \\ \midrule
        GPT     & 93 (0.17)* & 91 (0.21)* & \multirow{2}{*}{390} \\
     \ComFact   & 81 (0.05)  & 73 (0.04)  &                      \\ \midrule
        GPT     & 93 (0.10)  & 80 (0.16)* & \multirow{2}{*}{300} \\
     \Cicero    & 88 (0.05)  & 70 (0.06)  &                      \\ \midrule
        GPT     & 89 (0.08)  & 86 (0.08)  & \multirow{2}{*}{300} \\
     \Reflect   & 91 (0.09)  & 82 (0.04)  &                      \\ \bottomrule
\end{tabular}
}
\caption{The average \% ($\sigma<2\%$) of total samples (\textbf{\#}) tested as reasonable (\textbf{R}) and novel (\textbf{N}), with discordance probabilities in parentheses. \textbf{*}: statistical significance (McNemar's, $\alpha=0.05$). 90 more samples are used for \ComFact\ due to its greater number of inference types.}
\label{tab:quality_results}
\vspace{-1.5em}
\end{table}

\begin{figure*}[htb!]
    \centering
    \includegraphics[width=\textwidth]{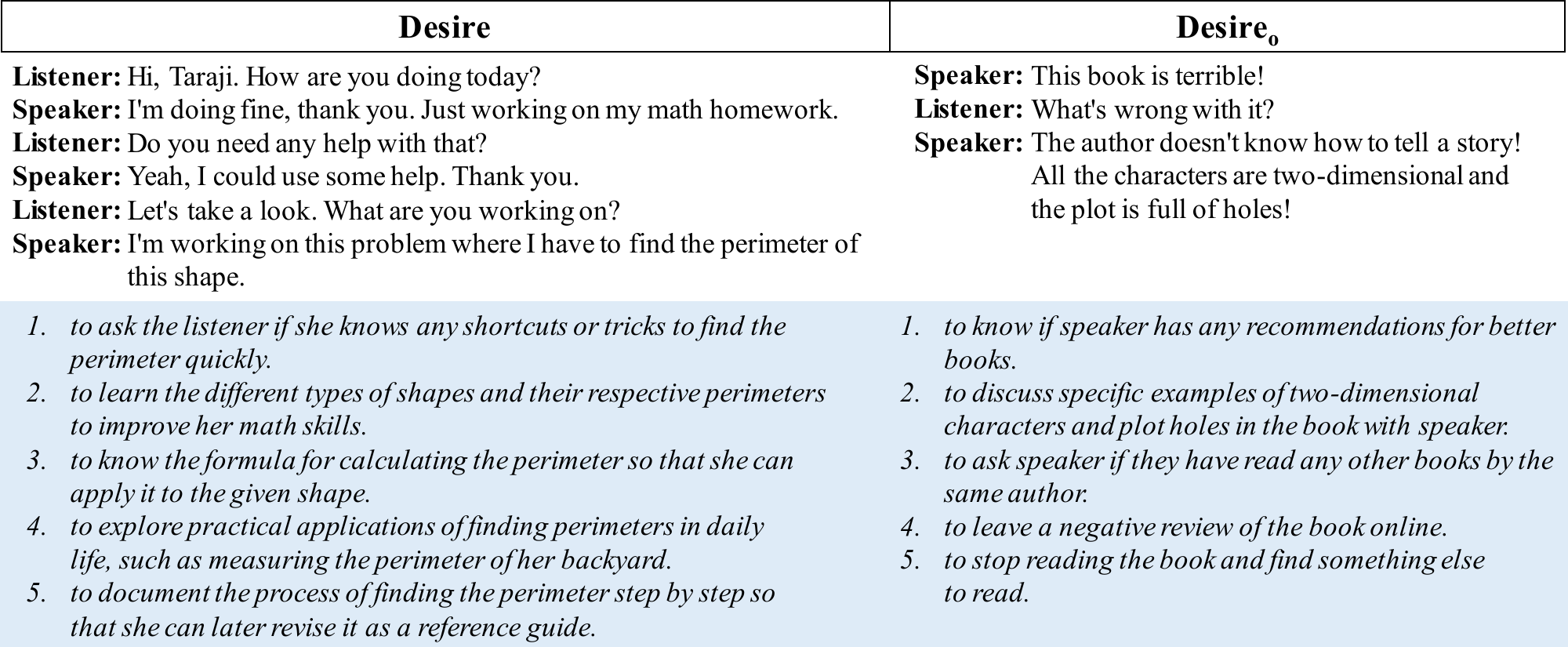}
    \vspace{-1.5em}
    \caption{Desire and Desire$_o$ inferences in the \ConvoSense\ dataset.}
    \label{fig:data_examples}
\end{figure*}



\section{ConvoSense Dataset}
\label{sec:convosense}

Given our assessment of high-quality, novel, and detailed GPT-generated commonsense inferences across various dialogue contexts and inference types (Section~\ref{sec:gpt_generation}), we construct a substantial conversational commonsense dataset using GPT, termed \ConvoSense.


\subsection{HumanGen: Human-generated Datasets}
\label{sec:existing_human_datasets}

For fair comparisons to our work, we combine the three human-generated datasets (Section~\ref{sec:related_work}) into a solitary dataset, termed \HumanGen.\footnote{Many commonsense types have a sparsity of training data when the human-generated datasets are viewed in isolation, which would impede the training of a neural model to adequately capture the commonsense type.}
Specifically, their train/validation/test sets are integrated independently.
For \ComFact\ and \Cicero, this integration follows the provided splits, while for \Reflect, data is sampled following an 80/10/10 distribution.
To standardize \HumanGen\ into a cohesive format, we perform the following preprocessing steps.
 
First, we leverage the mapping outlined in Table~\ref{tab:human-gen-datasets-types} along with the specifications from Table~\ref{tab:gpt-prompts} to identify relevant commonsense inference questions for each instance.
Then, we combine consecutive utterances from the same speaker to ensure every dialogue turn represents a distinct speaker.
Lastly, we apply \textit{Speaker} and \textit{Listener} tags in a similar manner to \ConvoSense\ (Figure~\ref{fig:data_examples}).
Since human-generated inferences often contain nominal references to specific target entities, we additionally incorporate the names of conversational participants into the tags, as exemplified by ``Speaker (A)''.

The naming conventions vary across the different human-generated datasets.
To maintain uniformity, we adopt the naming conventions used in \Cicero\ for both \ComFact\ and \Reflect, as \Cicero\ constitutes nearly 90\% of \HumanGen.
In \Cicero, participants are denoted as A and B. 
For \ComFact, originally lacking speaker designations, we randomly assign A/B tags to each conversation. 
On the other hand, \Reflect\ includes original speaker names; thus, we replace them with A/B tags accordingly.
Since the speaker name frequently appears in \Reflect's inferences, we uniformly replace it with ``\textit{the speaker}'', aligning with the prevalent format in \Cicero.

\subsection{ConvoSense: New GPT-generated Dataset}
\label{ssec:gpt-generated_dataset}

\begin{table*}[ht]
\centering
\resizebox{\textwidth}{!}{
\begin{tabular}{c|ccccc|cccc}
\toprule
 & \multicolumn{5}{c|}{\textbf{All}} & \multicolumn{4}{c}{\textbf{Poly}} \\
 & \multicolumn{1}{c}{\bf Examples} & \multicolumn{1}{c}{\bf Words} & \multicolumn{1}{c}{\bf Inferences} & \multicolumn{1}{c}{\bf U1(\#)} & \multicolumn{1}{c|}{\bf U2(\#)} & \multicolumn{1}{c}{\bf Examples} & \multicolumn{1}{c}{\bf U1(\%)} & \multicolumn{1}{c}{\bf U2(\%)} & \multicolumn{1}{c}{\bf UL(\%)} \\
\midrule
\bf ConvoSense &      120,000 &      14.6 & $\:\:$5.1 (2-13) &      16,666 &      199,087 &      120,000 & 92.8 & 98.9 & 98.8 \\ \midrule
\ComFact & $\;\;\;\;$3,909 & $\:\;$3.2 & $\:\:$1.4 (2-12) & $\quad\;$295 & $\quad\;\;\;$315 & $\:\;\:\;$1,401  & 86.7 & 97.3 & 60.3 \\
\Cicero & $\:\;$52,644 & 11.6 & $\:\:$1.3 (2-11) & $\:\;$7,598 & $\:\;$44,234 & $\:\;\:\;$9,911  & 84.4 & 97.2 & 98.7 \\
\Reflect & $\;\;\;$3,000 & $\:\;$6.4 & 1.1 (2-4) & $\quad\;$835 & $\:\;\:\;$1,407 & $\quad\;\;\;$216  & 85.1 & 95.2 & 82.2 \\
\midrule
\bf HumanGen   & $\:\;$59,553 & $\:\;$6.6 & $\:\:$1.3 (1-12) & $\:\;$2,886 & $\:\;$15,420 & $\:\;$11,528 & 86.7 & 97.0 & 78.3 \\
\bottomrule
\end{tabular}
}
\caption{Statistics of the \ConvoSense$\:$and \HumanGen$\:$datasets. \textbf{Poly}: polymorphic  examples (multiple inferences). \textbf{Examples}: \# of examples, \textbf{Words}: average \# of words per inference, \textbf{Inferences}: average \# of inferences per example with range shown in parentheses, \textbf{U1/2(\#)}: average \# of unique unigrams/bigrams across all inferences, \textbf{U1/2(\%)}: average \% of unique unigrams/bigrams between inferences within a single example, \textbf{UL(\%)}: average \% of unique inferences across all examples. Averages are calculated at the macro level across all inference types.}
\label{tab:data_stats}
\vspace{-1em}
\end{table*}

Constructing a practical dataset of commonsense inferences for dialogue benefits from covering a wide variety of dialogue situations. To this end, our construction process of \ConvoSense\ first carefully selects the dialogues to include based on their topical diversity, trims the dialogue contexts to optimize utterance diversity, and finally generates the inferences for each context.

\paragraph{Dialogue Selection} 

We choose to sample the dialogues for \ConvoSense\ as a subset of those dialogues in the high-quality and large-scale SODA dataset. SODA contains over a million dyadic dialogues generated by GPT covering situations based on ATOMIC commonsense tuples \cite{kim:22b}. For cost practicality, \ConvoSense\ is constructed to contain 10,000 training dialogues, 1,000 validation dialogues, and 1,000 test dialogues each. 

To encourage diversity in \ConvoSense, we employ BERTopic \cite{grootendorst:22}, which clusters the dialogues selected from SODA into groups using dimension reduction technique UMAP \cite{mcinnes:18} and HDBSCAN clustering algorithm \cite{mcinnes:17} on the BERT embeddings of the dialogues.\footnote{The \texttt{all-mpnet-base-v2} model is used for BERT.}
We configure the hyperparameters\footnote{{neighbors}: $5$, {components}: $5$, {min\_cluster\_size}: $2$.} to effectively group dialogues while maintaining a well-balanced distribution of group lengths based on manual verifications.
As a result, we obtain 100K dialogue groups, where each group consists of 6.3 dialogues on average. These groupings represent 100K unique dialogue topics, thus enabling the construction of \ConvoSense\ to span a variety of topics by sampling dialogues from a subset of these groupings.

Next, we randomly select one dialogue from the $n$ groupings, where each dialogue contains at least 5 utterances and has a BERTopic score of at least $0.95$ to its group.
To maintain distinct dialogue scenarios in each split, each grouping can only be selected for one split. 
Through this procedure, we set $n$ values as $[10K, 1K, 1K]$ for assembling the training, validation, and test splits, respectively.

\paragraph{Utterance Selection} 

For each selected dialogue, we determine which utterance to perform inference generation on.
We use the topic keywords identified for each group during the BERTopic grouping to pinpoint the most topically salient utterance in each dialogue and ensure that the diversity afforded by the grouping is maintained.
This is achieved by selecting the utterance whose embedding yields the highest cosine similarity with the embedding of the four-word topic string assigned to the dialogue's respective group by BERTopic.
Subsequently, we trim the dialogue's utterances such that the conversation ends at this selected utterance. 
This trimmed version becomes the final dialogue context used for commonsense inference generation, where the inferences are derived for the last utterance.


Because commonsense inferences often relate to a central figure in a conversation, either the speaker or the listener, we introduce nominal tags for the two participants. 
The terminal utterance is labeled as \textit{Speaker}, and its preceding utterance is labeled as \textit{Listener}.
These nominal tags are then assigned in alternating order to the remainder.  


\paragraph{Inference Types} 

For each preprocessed dialogue, GPT generates inferences for all included commonsense types following the procedure in Section~\ref{sec:gpt_generation}. 
Specifically, ten commonsense types are included: Subsequent, Cause, Prerequisite, Motivation, Attribute, Reaction, Reaction$_o$, Desire, Desire$_o$, and Constituents (highlighted in Table \ref{tab:gpt-prompts}). 
These types are selected based on their usage frequency in existing datasets and their lack of semantic overlap.


\paragraph{Data Statistics} 

Table~\ref{tab:data_stats} presents data statistics for\LN \ConvoSense\ and \HumanGen.
\ConvoSense\ significantly surpasses \HumanGen\ for data volume, particularly regarding instances with polymorphic outputs, where multiple inferences can be derived per instance.
Moreover, \ConvoSense\ boasts greater vocabulary diversity and reduced redundancy among inferences.
Illustrative examples from each dataset are shown in Figure \ref{fig:data_examples}.


\paragraph{Data Quality}

The results in Section \ref{sec:gpt_commonsense_results} demonstrate that GPT is generally capable of producing high-quality commonsense inferences regardless of the underlying dialogue source. Consequently, applying GPT to generate commonsense inferences for the SODA dialogues is expected to perform with similar high quality. To explicitly verify this, we conduct an evaluation of the \ConvoSense\ dataset. An external conversational AI expert, unaffiliated with this study, evaluates the generated inferences for 100 \ConvoSense\ examples (508 total inferences; average 5.08 inferences per example), with all ten inference types uniformly represented across examples. The human judge completes two evaluation tasks: grading reasonability and novelty of an inference (Sec.~\ref{sec:gpt_commonsense_metrics}) and performing inference clustering to measure per-example output diversity (Sec.~\ref{sec:auto_diversity}).  Table \ref{tab:convosense_eval} presents the results, confirming the high reasonability, novelty, detailedness, and diversity of the inferences in the \ConvoSense\ dataset.

\begin{table}[htbp]
    \centering
    \begin{tabular}{r|c}
    \toprule
    & \bf ConvoSense \\
    \midrule
     Reasonable    &  91\\
     Novel & 97 \\
     Detailed & 63 \\
     Clusters & 4.82 (95\%) \\
    \end{tabular}
    \caption{Human evaluation results on 100 examples of ConvoSense data, including the \% of total inferences judged to be reasonable and novel, the \% of positive novelty inferences judged to be detailed (vs. simple), and the average number of unique inference clusters per example, with the average \% of unique inferences per example in parentheses.}
    \label{tab:convosense_eval}
    \vspace{-1em}
\end{table}

\paragraph{Error Analysis}

We next perform an error analysis on the unreasonable inferences identified by the human judge. We observe that most unreasonable inferences are explained by being too niche to be likely given only the provided information in the dialogue context (26\%; Desire examples \#4-5 in Figure \ref{fig:data_examples}), or by their attribution to the wrong conversational participant (26\%; Desire$_o$ examples \#4-5 in Figure \ref{fig:data_examples}). Relatively speaking, only a small percentage of unreasonable inferences are explained by a violation of common knowledge of human experiences (10\%), a lack of relevance to the dialogue context (10\%), or a contradiction of the dialogue context (7\%). This suggests that \ConvoSense\ inferences are predominately accurate representations of commonsense understanding, although they can suffer from lack of precision regarding situational nuances and speaker roles.

\section{Generative Commonsense Models}
\label{sec:model_strategies}

\subsection{Training and Decoding Strategies}

With the rich and diverse multi-inference examples provided in \ConvoSense, we are well-positioned for training commonsense generation models that\LN produce versatile outputs. 
Yet, a key query remains: how can we induce this versatility into the model? 

A common method of enhancing diversity in generative outputs is to modify the decoding strategy\LN \cite{gimpel:13, vijayakumar:16, ippolito:19}. 
Through preliminary testing, we observe that diverse beam search decoding with Hamming distance reward following \citet{vijayakumar:16} improves the output diversity with less impact on accuracy compared to other methods. 

On the other hand, \citet{cao:20} propose modifying the model architecture by introducing latent variables to guide output variety.
However, these approaches only approximate learning varied responses by relying on conditioning on random latent variables. In contrast, \ConvoSense\ provides direct access to numerous inferences per input, enabling direct training of generative models that produce multiple inferences per example, with the set of inferences treated as target outputs during training. 
Therefore, we explore the performance of three strategies for diverse generation of commonsense inferences.


\paragraph{Monomorphic Beam Search (M)}

This model receives as input a dialogue context $\bm{C}$ consisting of\ the previous six utterances delimited by their corresponding speaker tags, the current response $\bm{r}$ for which to generate inferences, and a commonsense question $\bm{q}$ pertaining to one of the ten inference types (Table~\ref{tab:gpt-prompts}) in the following format: 
\begin{center}
$\bm{C}\backslash n \mathbf{r}\backslash n\backslash n\mathrm{[Question]}\bm{q}\backslash n\mathrm{[Answer]}$
\end{center}
It is trained to output a single inference~$i$. During training, instances with multiple correct inferences $I$ generate several training examples, one for each target inference $i \in I$. 
During inference, standard beam search decoding is used to generate \textit{k} outputs.


\paragraph{Monomorphic Diverse Beam Search (M*)}

This model adheres to the same design as the M model, except during inference, it uses Hamming-distance diverse beam search decoding instead to generate \textit{k} outputs, following \citet{vijayakumar:16}.


\paragraph{Polymorphic (P)}

Using the same input as the M model, this model is trained to output a series of\LN inferences as a sequence. 
To do this, the ground-truth inferences for each training example are concatenated into a list-like sequence, delimited by semicolons and prefixed by an integer representing their position in the list as follow: 
\begin{center}
(1) $\bm{i_1}$; (2) $\bm{i_2}$; (3) $\bm{i_3}$; $\ldots$
\end{center}
The order of the answers in the list are shuffled between each training epoch. 
During inference, standard beam search decoding is used to generate the top-1 output. 
A single output from this model is intended to represent the set of multiple diverse inferences for the input, without the need for any post-hoc decoding strategies, which other studies have observed to negatively impact the accuracy of the output generations \cite{ippolito:19}.


\subsection{Model Configuration}
\label{ssec:model_configuration}

We develop six generative models: {ConvoSenseM}, {ConvoSenseM*}, {ConvoSenseP}, {HumanGenM}, {HumanGenM*}, and {HumanGenP}. 
Each model name denotes the training dataset with the terminal letter indicating the model strategy.
All of them use {T5-3b} \cite{raffel:20} as the base model, which is then finetuned on the corresponding dataset following the indicated model strategy. 
The ConvoSense* and HumanGen* models are finetuned for $5$ or $10$ epochs, respectively.
The best-performing models\LN and hyperparameters\footnote{The Adafactor optimizer is used with a weight decay of 5e-3 and a learning rate of 5e-6, except for ConvoSenseP with 1e-6. The max source length is set to 768. The max target length is set to 400 for P models and 128 for other models. All models are trained using bf16 for memory efficiency. P models use a prefix of ``provide several reasonable answers to the question based on the dialogue:$\backslash$n'' and other models use a prefix of ``provide a reasonable answer to the question based on the dialogue:$\backslash$n''.} are selected through grid-search based on their results on the validation sets.




For all models, decoding is performed with 10 beams.
For ConvoSenseM* and HumanGenM*, the number of beam groups is $10$, and the diversity penalty is $0.5$ and $1.0$, respectively.
For P models, decoding also uses a repetition penalty of $5.0$ to reduce output token repetition.


It is worth noting that only 16\% of HumanGen examples feature multiple ground-truth inferences. 
Training a P model on the complete dataset yields a single-inference model, which defeats the purpose of the \textit{polymorphic} model strategy. 
Instead, we develop the HumanGenP model exclusively on multi-inference instances to facilitate learning of polymorphic outputs.

\begin{table*}[htbp!]
\centering\resizebox{\textwidth}{!}{

\begin{tabular}{l|ccc|ccc|ccc|ccc}
\toprule
& \multicolumn{6}{c|}{\cellcolor{gray!20}\textbf{HumanGen Test Split ($\bm{n=11,494}$)}} & \multicolumn{6}{c}{\cellcolor{gray!20}\textbf{ConvoSense Test Split ($\bm{n=10,000}$)}} \\
& \multicolumn{3}{c|}{\textbf{Top-1}} & \multicolumn{3}{c|}{\textbf{Top-5}} & \multicolumn{3}{c|}{\textbf{Top-1}} & \multicolumn{3}{c}{\textbf{Top-5}} \\
& \bf BLEU & \bf BS & \bf Embed & \bf BLEU & \bf BS & \bf Embed & \bf BLEU & \bf BS & \bf Embed & \bf BLEU & \bf BS & \bf Embed \\
\midrule
ConvoSenseM & \underline{5.407} & \underline{0.641} & \underline{0.422}$^\dagger$ & 6.282 & 0.650 & 0.462 & \underline{19.019} & \underline{0.777} & \underline{0.730}$^\dagger$ & 11.119 & 0.700 & 0.603 \\
ConvoSenseM* & 5.131 & 0.637 & 0.416 & \underline{6.710} & \underline{0.658}$^\dagger$ & \underline{0.496} & 17.923 & 0.773 & 0.725 & \underline{11.933} & \underline{0.709} & \underline{0.627} \\
ConvoSenseP & 4.922 & 0.635 & \underline{0.422} & 6.026 & 0.645 & 0.482 & 15.163 & 0.758 & 0.703 & 9.725 & 0.644 & 0.564 \\
\midrule
HumanGenM & \underline{10.724} & \underline{0.711} & \underline{0.538} & \underline{12.701} & 0.721 & 0.576 & 5.095 & 0.633 & 0.501 & \underline{3.574} & \underline{0.588} & \underline{0.413}  \\
HumanGenM* & 9.473 & 0.697 & 0.511 & 12.056 & \underline{0.724}$^\dagger$ & \underline{0.591} & 4.263 & 0.617 & 0.481 & 3.045 & 0.571 & 0.393  \\
HumanGenP & 9.524 & 0.700 & 0.523 & 9.658 & 0.645 & 0.504 & \underline{6.358} & \underline{0.655} & \underline{0.528} & 2.330 & 0.256 & 0.201  \\
\bottomrule
\end{tabular}
}
\caption{Reference metric results on test splits. Columns \textbf{BS} denote Bertscore. \underline{Underline} indicates best metric with statistical significance under Bonferonni multi-test correction, except where indicated by $\dagger$ (t-test, $\alpha=0.05$).}
\label{tab:reference_eval_results}
\vspace{-1em}
\end{table*}

\section{Generative Model Evaluation}
\label{sec:model-evaluation}

We evaluate the six generative models (Section~\ref{ssec:model_configuration}) on the ten commonsense inference types (Table~\ref{tab:gpt-prompts}) that exist in both the \HumanGen\ (Section~\ref{sec:existing_human_datasets}) and \ConvoSense\ (Section~\ref{ssec:gpt-generated_dataset}) datasets.
The model performance is evaluated using automatic reference metrics (Section~\ref{ssec:reference_metrics}), automatic diversity metrics (Section~\ref{sec:auto_diversity}), and human evaluations of reasonability and novelty (Section~\ref{ssec:human_evaluations}).

\subsection{Automatic Reference Metrics}
\label{ssec:reference_metrics}

Conventional evaluations of generative models against ground-truth references often overlook the diverse nature of the outputs. They typically assess individual model outputs against a single reference, focusing on best-case performance due to dataset constraints. However, such assessments are inadequate for our multi-inference dialogue generation objective. To address this, we structure our automated evaluation method to account for the concept of output diversity. This method, referred to as \texttt{PolyAgg}, serves as an aggregation function compatible with standard evaluation metrics. Its purpose is to gauge the model's capacity to encompass the complete set of ground-truth references in its generated outputs.

\begin{algorithm}
\small 
\caption{Metric Aggregation}
\begin{algorithmic}[1]
\Procedure{PolyAgg}{$outputs$, $references$}
    \State $matrix \gets []$
    \For{$o \in outputs$}
        \State $row \gets []$
        \For{$r \in references$}
            \State $score \gets \Call{Metric}{o, r}$
            \State $\Call{Append}{score, row}$
        \EndFor
        \State $\Call{Append}{row, matrix}$
    \EndFor
    \State $a \gets \Call{LinearSumAssignment}{matrix}$
    \State \Return $\Call{Mean}{a}$
\EndProcedure
\end{algorithmic}
\end{algorithm}

\noindent Algorithm 1 demonstrates the \texttt{PolyAgg} aggregation function. It computes a score matrix for each example, where rows represent model outputs and columns represent ground-truth references, and finds the maximal assignment of rows to columns following the linear sum assignment problem \cite{burkard:99}, which seeks to find the optimal bijective mapping between rows and columns in a cost matrix. By mandating a one-to-one mapping from model outputs to references, we can accurately measure reference set coverage and prevent models that generate mere surface-level variations from scoring highly on datasets with diverse references. We use SciPy's linear sum assignment solver, then calculate the mean of the assigned scores for the final metric value. \citet{dou:21} utilize a similar aggregation for evaluating a diverse dialogue response generation model.

One consideration for \texttt{PolyAgg} is that it can only match up to the number of generated outputs. If a model generates fewer outputs than there are references, \texttt{PolyAgg} will not measure against all references. However, this is a reflection of the model's coverage capability, which is valuable information. To capture this, we introduce a coverage moderator for the \texttt{PolyAgg} score. Using cardinality notation $|\cdot|$, where $outs_e$ denotes the model outputs and $refs_e$ denotes the ground-truth references for a single example $e \in E$, the coverage moderator \texttt{C} is defined as:

\vspace{-0.5em}
\begin{equation}
    C = \frac{|outs_e|}{| refs_e |}
\end{equation}
\vspace{-0.5em}

\noindent Furthermore, different dialogue contexts can vary in the amount of diversity to their inferences, due to the nature of the described situations or shared information within the dialogue. A model achieving a high \texttt{PolyAgg} score on a diverse example should receive greater reward compared to a low-diversity case. Thus, not all examples should be treated equally when computing the overall model score; rather, each score should be proportionally weighted based on the corresponding number of ground-truth references.

Combining the \texttt{PolyAgg} aggregation, coverage moderator \texttt{C}, and diversity weighting, the final score for a model is calculated as:

\vspace{-0.5em}
\begin{equation}
\label{eq:polymorphic_metric}
    \frac{\sum\limits_{e \in E} PolyAgg(outs_{e}, refs_{e}) * C * |refs_e|}{\sum\limits_{e \in E}{|refs_{e}|}}
\end{equation}
\vspace{-1.2em}

\noindent We use this evaluation scheme with three automatic metrics to measure the performance of the models. We include the traditional ngram-matching BLEU metric with $n \in [1,4]$ \cite{papineni:02}, the embedding-based metrics BertScore\footnote{BertScore: \texttt{microsoft/deberta-xlarge-mnli}} \cite{zhang:19}, and sentence cosine similarity using SentenceBert\footnote{SentenceBert: \texttt{all-mpnet-base-v2}} \cite{reimers:19}.


\paragraph{Results}

We evaluate each model in terms of both its best-case performance (Top-1 output) and its multi-inference performance (Top-5 outputs). In the Top-1 setting, the maximum score achieved by the top-1 output against all of the ground-truth references for an example is taken and averaged across the test data. In the Top-5 setting, the top-5 outputs from the models are taken and scores are calculated using Equation \ref{eq:polymorphic_metric}, before being averaged across the test data. For M(*) models, the top one or five beams are taken as the outputs for each setting. For P models, the first one or five inferences in the outputted sequence are taken as the outputs for each setting. The results are shown in Table~\ref{tab:reference_eval_results} for each model on the \HumanGen\ and \ConvoSense\ test splits, respectively. 

Overall, it is evident that using diversity-promoting decoding (M*) outperforms the direct generation of multiple inferences (P). 
This approach achieves the highest BLEU, BertScore, and sentence similarity scores in the Top-5 assessment setting. This trend is particularly pronounced in the case of the ConvoSense-trained model, holding true for both the \ConvoSense\ and \HumanGen\ test splits. Enhancing training inference diversity as seen in \ConvoSense\ appears to support the adoption of diversity-focused decoding strategies, yielding more contextually relevant outputs aligned with ground-truth references, even when applied to test examples from different datasets.

In the Top-1 setting, monomorphic models with standard beam search demonstrate superior performance for both HumanGen- and ConvoSense-trained models. However, the difference compared to diverse beam search is relatively minor, particularly when considering embedding-based metrics. Interestingly, the HumanGenP model displays the strongest ability to generalize to the \ConvoSense\ test split among all HumanGen-trained models in the Top-1 scenario. Upon manual comparison of HumanGenP outputs against other HumanGen-trained models, we observe that HumanGenP is more inclined to specify a focal person in the inference (e.g., "the speaker/listener"). This often aligns better with \ConvoSense\ references, although in a superficial manner with little impact on the underlying semantics.

It is also observed that the models produce low scores when evaluated against the test examples that are out-of-distribution with respect to their training data. This may not reflect the true underlying reasonability of the generated inferences, but rather a difference in inference content between the datasets, which is supported by evidence in Section \ref{sec:gpt_commonsense_results} showing that human-written generations are more often repetitive with the dialogue context than GPT generations. To obtain a direct measure of the quality of the generated model inferences, we perform a human evaluation in Section \ref{ssec:human_evaluations}. 



\subsection{Automatic Diversity Metrics}
\label{sec:auto_diversity}

To assess the ability of each model in generating diverse inferences for a given dialogue context, we employ a clustering approach under the Top-5 evaluation scheme. This involves grouping the model generations for each example into clusters of inferences with similar meanings. The average number of inference clusters across examples serves as a measure of output diversity.

For each of the ten inference types, we draw 50 examples from the test splits of \ConvoSense\ and \HumanGen, except for the \texttt{Constituents} type in \HumanGen\ due to its smaller test split (22 examples). We instruct GPT4\footnote{\texttt{gpt-4-0613} with a temperature setting of 0} to create groups of semantically similar inferences given a dialogue context, question, and a list of inferences. GPT4 demonstrates its proficiency by achieving an average B-cubed F1-score \cite{bagga:98} of 0.872 against clusters identified by one of the authors for 20 examples, where B-cubed is a common clustering evaluation metric that measures the precision and recall of each element's neighbors within the same cluster. This outperforms Amazon Mechanical Turk crowdworkers who only achieved a score of 0.581.\footnote{The self-serve SurgeAI crowdsourcing platform previously used in Section \ref{sec:gpt_commonsense_metrics} was discontinued during this work.}

\paragraph{Results}

Table \ref{tab:diversity_eval_results} displays diversity outcomes per model. 
For both HumanGen and ConvoSense-trained models, the monomorphic model with diverse beam search generates the most unique outputs.\footnote{High unique percentages for P models are due to low-count inference output (average of 4.4 and 2.0 outputted inferences for ConvoSenseP and HumanGenP, respectively).}
While ConvoSenseM* slightly outperforms HumanGenM* in terms of inference diversity, both models exhibit similar unique inference cluster counts. Compared to the \ConvoSense\ inferences themselves (Table~\ref{tab:convosense_eval}), it is clear that none of the trained models are able to replicate the high inference diversity.
Nonetheless, there is a large discrepancy in inference detail, which is revealed through human assessments in the next section. 

\begin{table}[htp!]
\centering
\small{ 
\begin{tabular}{l|c|c}
\toprule 
& \bf Clusters & \bf Words \\
\midrule 
ConvoSenseM & 2.680 (54\%) & 12.179 \\

ConvoSenseM* & \underline{3.509} (70\%) & 12.928 \\

ConvoSenseP & 3.262 (74\%) & 13.292 \\
\midrule 
HumanGenM & 3.031 (61\%) & 6.492 \\

HumanGenM* & \underline{3.452} (69\%) & 5.544 \\

HumanGenP & 1.348 (69\%) & 7.744 \\
\bottomrule 
\end{tabular}
}
\caption{Diversity metric results. \textbf{Clusters}: average inference clusters identified per example (with average \% of unique inferences per example in parentheses). \underline{Underline} indicates statistical significance in number of clusters within-block (t-test, $\alpha=0.05$). Cross-block (ConvoSenseM* vs HumanGenM*) significance is not achieved.  \textbf{Words}: average number of inference words.}
\label{tab:diversity_eval_results}
\vspace{-1.5em}
\end{table}






\subsection{Human Evaluations}
\label{ssec:human_evaluations}

We also evaluate the models through human assessment, in both the Top-1 and Top-5 setting. Based on automated evaluation outcomes, we compare ConvoSenseM* to both HumanGenM and HumanGenM*. An external conversational AI expert, unaffiliated with this study, evaluates the top five inferences for 60 examples per model in a blinded design, with all ten inference types and both datasets being uniformly represented. The human judge completes two evaluation tasks: grading reasonability and novelty of an inference (Sec. \ref{sec:gpt_commonsense_metrics}) and performing inference clustering (Sec. \ref{sec:auto_diversity}). 

\begin{table}[htbp!]
\centering
\small{ 
\begin{tabular}{l|cc|ccc}
\toprule
& \multicolumn{2}{c|}{\textbf{Top-1}} & \multicolumn{3}{c}{\textbf{Top-5}} \\
 & \bf R & \bf N & \bf R  & \bf N & \bf Clusters \\
\midrule
ConvoSenseM* & \underline{90} & \underline{98} & \underline{93} & \underline{98} & 3.42 (68\%) \\
HumanGenM    & 75 & 57 & 81 & 56 & 2.25 (45\%) \\  
HumanGenM*   & 75 & 70 & 81 & 70 & 3.17 (63\%) \\
\bottomrule
\end{tabular}
}
\caption{Percentage of reasonable (\textbf{R}) and novel (\textbf{N}) inferences from each model. \underline{Underline} denotes a statistically significant result against both HumanGen models (chi-square proportions test, $\alpha=0.05$). The average number of inference clusters is also shown, along with the average \% of unique inferences per example in parentheses (\textbf{Clusters}). }
\label{tab:model_human_eval}
\vspace{-1em}
\end{table}

\paragraph{Results} Table \ref{tab:model_human_eval} demonstrates ConvoSenseM*'s superior performance compared to the HumanGen models. ConvoSenseM* achieves a remarkable 93\% reasonability and 98\% novelty, averaging 3.4 unique inferences per example. Indeed, similar results hold even when considering the Top-1 output per model, showing that ConvoSenseM* exhibits strong performance regardless of whether a single-best inference is desired or a diverse set of inferences are desired. 
Moreover, when considering the positive novelty inferences in the Top-5 setting, we observe that 75\% are annotated as \textit{detailed} for ConvoSenseM* whereas only 7\% are indicated as such for HumanGenM*. This reveals a substantial improvement in the amount of detail present in the inferences produced by ConvoSense models as compared to HumanGen models, which results in richer information being provided by the model.

\section{Limitations and Ethical Considerations}

This work does not intend to present an exhaustive set of commonsense inferences for dialogue. While we adhere to established inference types relevant to dialogue from existing literature, there could be overlooked types or unique challenges within specific dialogue domains that remain to be explored. 

Furthermore, it is important to recognize that some social commonsense inference types may be associated with stereotypes and biases. When employing a model that produces commonsense inferences in a setting that impacts human users, caution must be exercised to prevent unjust or prejudiced decisions. Although exploration of the prevalence of harmful biases is out of the scope of the current work, we welcome future investigations into quantifying these aspects of our resources.

Finally, we adhered to OpenAI's terms of service and related policies when utilizing GPT, and we acknowledge that any subsequent utilization of our models and data should refer to these policies.
\section{Future Work}

Although \ConvoSense\ is composed of diverse multi-inference dialogue data (Table~\ref{tab:convosense_eval}), it is clear from our experiments (Tables \ref{tab:diversity_eval_results} and \ref{tab:model_human_eval}) that our trained models do not quite achieve the same degree of inference diversity. Further work is needed on improving the ability of distilled models to better capture the diversity present in the data.

In addition, the integration of commonsense understanding into dialogue applications has shown promising results in improving performance on tasks such as response generation, summarization, and reading comprehension in previous works.
In light of this, our work on improving commonsense resources and models presents an opportunity for further advancements in these dialogue applications. In particular, future work exploring how to capitalize on our commonsense model for dialogue response generation is highly compelling, since commonsense errors are one of the most common issues for modern dialogue agents \cite{finch:23}. However, previous works have revealed that naive integration of commonsense inferences into neural models do not necessarily produce improvements \cite{zhou:22a}. As a result, we leave the integration of our commonsense model to future work to allow for thorough investigation of its impact on response generation, covering aspects such as the impact of different commonsense inference types, the filtering of relevant inference types per dialogue context, and the effect of synthesizing multiple inferences into dialogue responses.

\section{Conclusion}

In this work, we present \ConvoSense, an automatically constructed dataset of multi-output commonsense inferences for dialogue. \ConvoSense $\:$ surpasses existing datasets in size, \LN advances inference detail and novelty, and attains comparable (if not superior) reasonability when compared to existing datasets. Our investigation into various techniques for generating multiple inferences reveals that diverse beam search on single-output generative models yields the best outcomes. By publicly releasing our trained models, we enable other works to benefit from the remarkable improvements in commonsense reasonability and novelty achieved by this work. 

\section*{Acknowledgements}

We gratefully acknowledge the support of Amazon for this work. Any opinions, findings, and conclusions or recommendations expressed in this material are those of the authors and do not necessarily reflect the views of Amazon.
We would also like to thank the anonymous reviewers and the Action Editor for their valuable feedback.

\bibliography{custom}
\bibliographystyle{acl_natbib}

\end{document}